\documentclass{article} 
\usepackage{iclr2016_workshop,times}
\usepackage{hyperref}
\usepackage{url}
\usepackage{amsmath}
\usepackage{amssymb}
\usepackage{graphicx}
\usepackage{caption}
\usepackage{subfigure}
\usepackage{amsthm}

\usepackage{xcolor}
\newcommand{\mycomment}[3]%
{\par\noindent\fbox{\color{#1}\parbox{.99\columnwidth}{\textbf{#2}\ #3}}}

\title{A metric learning approach for graph-based label propagation}
\author{Pauline Wauquier\\
Clic And Walk\\
25 rue Corneille\\
59100 Roubaix\\
France\\
\texttt{pauline.wauquier@inria.fr} \\
\AND
Mikaela Keller\\
Univ. Lille, CNRS, Centrale Lille, UMR 9189 - CRIStAL\\
F-59000 Lille\\
France\\
\texttt{mikaela.keller@univ-lille3.fr}
}

\date{}
\begin{document}

\maketitle

\begin{abstract}
The efficiency of graph-based semi-supervised algorithms depends on the graph
of instances on which they are applied. The instances are often in a
vectorial form before a graph linking them is built. The construction of
the graph relies on a metric over the vectorial space that help define the
weight of the connection between entities. The classic choice for this
metric is usually a distance measure or a similarity measure based on the
euclidean norm. We claim that in some cases the euclidean norm on the initial vectorial
space might not be the more appropriate to solve the task efficiently. We
propose an algorithm that aims at learning the most appropriate vectorial
representation for building a graph on which the task at hand is solved
efficiently.

\end{abstract}

\section{Introduction}

Transductive or semi-supervised graph-based learning algorithms, such
as \cite{ZhuGL03, NIPS2003ZhouWeston, zhu05survey, liu2009robust}, take
advantage of both annotated and unannotate data to solve automated labeling
tasks. It relies on the homophily property: entities that are close in the
graph are supposed to have similar behaviors.


However, those graph-based learning algorithms are dependent on the
graph on which they are applied. Indeed the effectiveness of graph-based
learning algorithms relies on the relevance of the data representation for
the targeted task (\cite{MaierHeinLuxburgInfluence,JebWanCha09,SousaRB13}).

In general the graph representation of the data is built in two
steps \cite{liu2009robust,LuxburgGConstruction, SousaRB13}. In this
representation, every data point is seen as a vertex of the graph and the
first building step is to compute the edge weight between every pair of
vertices. This weight, refleting the similarity degree between two data
points is usually computed as a function of the euclidean scalar product
between the data points.
In the second step of the graph construction edge with low similarity
degree are discarded. This is done by applying a non-linear transformation
on the edges, like selecting a threshold $\epsilon$ on the weight value (a
method referred as $\epsilon$-graph), a fixed number $k$ of neighbors for
each node ($k$-nn graph) or by applying a kernel on the weights (making
high weights higher and low weights lower).

We claim that there might be cases where the euclidean space in
which the data points lie will not produce an optimal graph for solving the
targeted task.

To answer this concern, we propose an approach to build a graph from our
data but also adapted to a specific labeling task.
We learn a new representation of our data based on constraints related to
the task: Data points should be close in the graph if they share similar
labels. As the data may be too complex to be projected with a linear
approach, we will use a deep neural network in order to learn the most
appropriate representation of our data for the task.


Previous work related to ours fall into two main categories: representation
learning algorithms and metric learning approaches. Among representation
learning algorithms, some attempt at learning the best representation for a
supervised task in a semi-supervised setting. In particular, the work
presented in \cite{Chopra:LSMD, RifaiDVBM11, embedding, HofferICLR15} is
very close to our own work.
The main difference of \cite{RifaiDVBM11, embedding, HofferICLR15} with our
approach is that in those models the classifier is parametric while we rely
on a non-parametric classifier on which we give guarantees. Our main
difference \cite{Chopra:LSMD} approach, is the shape of their
representation function (convolutionnal network vs multi-layer perceptron)
and their exact learning criterion (pairwise comparison vs relative
comparison).

Among the metric learning approaches, the more popular ones have as
objective the learning of a linear re-weighting of the euclidean distance,
or Mahalanobis distance (for example \cite{LMNN, DhillonMISC}).
Although linear metrics are convenient to optimize, they are not able to
capture the non-linear structure of the data; some non-linear metric
learning algorithms have been developed and compose a second group of
metric learning approaches. Most non-linear metric learning are kernelized
version of linear metrics learning approaches (\cite{NIPS2012_1223, kdml})
and present the drawback of the choice of the kernel. In \cite{kernel}, a
deformed kernel is learn depending on the data geometry which will be used
to the classification task resolve.

Our approach stands in between metric learning and representation
learning. It focuses on a specific existing metric $w$ but learns to
project the data in a space in which $w$ is meaningful for the targeted
task.

\section{Algorithm's description}
\label{algo}

Let us define $D=\{(x_{i},y_{i})\}_{i=1..n}$ a dataset of examples such
that $\forall i \in \{1,...,n\}, x_{i}\in \mathbb{R}^{d}$ and
$y_{i} \in \{1,...,c\}$. Let $X$ be the projection of $D$ on
$\mathbb{R}^{d}$ and $C_{1},...,C_{c}$ be a partition of $X$, such that
$\forall k\in \{1,...,c\}$, $C_{k}=\{x_{i}\in X|(x_{i},y_{i})\in D \wedge
y_{i}=k\}$ and $\forall h\neq k$, $C_{k}\cap C_{h}=\varnothing$.

Let us assume that $D$ is partioned into a training set $D_{train}$ and a
test set $D_{test}$; we can define $X_{l}$ to be the projection of
$D_{train}$ on $\mathbb{R}^d$ and $X_{u}$ the projection of $D_{test}$ on
$\mathbb{R}^d$, $X=X_{l}\cup X_{u}$ . Labels of elements $x\in X_{u}$ are
hidden. If $T$ is the set of triplet constraints defined as:
$$T=\{(x,x_{+},x_{-})|(x,y),(x_{+},y_{+}),(x_{-},y_{-})\in D \wedge
y=y_{+} \neq y_{-}\},$$
then let us define $T = T_{train}\cup T_{test}$ such that
$T_{train}=\{(x,x_{+},x_{-})\in T|x,x_{+},x_{-}\in X_{l}\}$ and
$T_{test}=T\backslash T_{train}$.

Let $\phi:\mathbb{R}^d\leftarrow\mathbb{R}^q$ be a non-linear function.
Let us consider a distance measure $d$\footnote{Concepts introduced in the
following can be made true for a similarity measure instead, with a simple
inversion in the constraints inequality and in the cost function.} on
$\mathbb{R}^q$. We want the projection $\phi$ to respect the following
constraint with respect to distance $d$:
$$\forall (x,x_{+},x_{-})\in T, \>
d(\phi(x),\phi(x_{+}))<d(\phi(x),\phi(x_{-})).$$

At training time, and using a hinge loss we can reformulate our set of
constraints as the following cost function to minimize:
$$ C(\phi|T_{train}) = \underset{(x,x_{+},x_{-}) \in T_{train}}{\sum} max(0,1-[d(\phi(x),\phi(x_{-})-d(\phi(x),\phi(x_{+}))])$$

The objective function of our problem is:
$$ \phi = \arg \min_{\tilde{\phi}} C(\tilde{\phi}|T_{train})$$

In order to learn $\phi$, we train a \textit{Siamese} neural network \cite{}
containing three replicated non-linear neural network $\phi$
(Fig.~\ref{nnSiamese}), by stochastic gradient descent optimization.


Let us define $W \in \mathbb{R}^{n\times n}$ as the matrix whose components
$W_{ij}=d(\phi(x_{i}),\phi(x_{j}))$. $W$ can be seen as the adjacency
matrix of a complete graph.  We can use the computed graph in order to
predict the hidden label, through the well known label propagation
algorithm (\cite{Zhu02learningfrom,NIPS2003ZhouWeston}); In order to remove
the non relevant edges of the graph, a pruning phase is usually performed
on the graph. Different pruning methods can be applied. The creation of a
$k-nn$ graph, where only the edges for the $k$ nearest neighbors of each
instances are kept, is a popular one. Another pruning technique is the
extraction of the $\epsilon$-graph, where edges are kept depending on a
threshold $\epsilon$ on their weights values. The obtained matrix $W$,
pruned or not, is then row-normalized.

At each step of the label propagation algorithm, the label of an instance
 is decided according to the labels of its neighbors. Let us define
 $F^{m}\in\mathbb{R}^{n\times c}$ such that $F^{m}_{ik}$ is the probability
 at iteration $m$ for $x_{i}$ to belong to the class $C_{k}$; let's define
 $F^{0}_{ik} = \begin{cases} 1 &\mbox{ if } x_{i} \in X_{l}\cap C_{k} \\ 0
 &\mbox{ otherwise} \end{cases}$.  Let's consider the clamped label
 propagation algorithm, i.e. the label of training instances are not
 modified during the different epochs:
$F^{m+1}_{ic} = \begin{cases}
  (WF^{m})_{ic} &\mbox{ if } x_{i}\in X_{u}\\
  F^{0}_{im} &\mbox{ if } x_{i} \in X_{l} \\
\end{cases}$.
Let $LP(x_{i})$ be the label predicted by the label propagation algorithm
for $x_{i} \in X_{u}$. The label is computed as $LP(x_{i})
= \text{arg } \underset{k}{\text{max }} F^{\infty}_{ik}$.

\section{Theoretical guarantees}
\label{proof}
We claim that under some initial assumptions, the algorithm described in
section \ref{algo} provides a graph representation of the data that is
optimal for classification through the label propagation algorithm. In the
following we substantiate our claim by showing that we can find an
$\epsilon$ value for pruning the complete graph obtained from our data in
the projected space in such a way that the resulting $\epsilon$-graph is
made of $c$ connected components homogeneous in class. 
We show that we can ensure the existence of such an $\epsilon$ if we
suppose that each testing triplet is in a close neighbourhood of at least
one similarly labeled training triplet.

We prove our claim through three main steps. We first state that we can
bound the distance between the projection of points depending on their
distance in the initial space.
In a second step we show that if a triplet is properly projected in the new
space, then triplets that are projected nearby are also properly projected.
By bringing together the first two steps we are able to show that triplets
that are close in the initial space to a triplet that is properly projected
is also projected properly.
Knowing that, we can finally exhibit an $\epsilon$ that allows us to
compute an optimal $\epsilon$-graph for the label propagation algorithm.


Let the distance be the euclidean distance and our transformation be a
multi-layered perceptrons
$\tilde{\phi}, \phi: \mathbb{R}^{d} \mapsto \mathbb{R}^{q}$, with one
hidden layer, as

$$\tilde{\phi}(x)=\tilde{B}\tanh(\tilde{A}x+\tilde{\alpha})+\tilde{\beta}
 \>\textrm{ and }\>\phi(x)=B\tanh(Ax+\alpha)+\beta,$$ where $\tilde{B},
 B\in \mathbb{R}^{q\times p}$, $\tilde{A}, A\in \mathbb{R}^{p\times d}$,
 $\tilde{\alpha}, \alpha\in \mathbb{R}^{p}$ and
 $\tilde{\beta}, \beta\in \mathbb{R}^{q}$.  Let's suppose that $B$, $A$,
 $\alpha$ and $\beta$ are learned, with no error, such that
$$\phi = \mbox{arg }\underset{\tilde{\phi}}{\mbox{min }} \underset{(x,x_{+},x_{-})\in T_{train}}{\sum} \mbox{max }[0,1-d(\tilde{\phi}(x),\tilde{\phi}(x_{-}))+d(\tilde{\phi}(x),\tilde{\phi}(x_{+}))]$$ 

Based on $\phi$, our first lemma claims that we can easily bound the
euclidean distance between the projection of two points by a factor of
their initial euclidean distance, where the factor is dependent on $A$ and
$B$. This can easily be done by bounding the application of each
transformation of $\phi$.

Our second lemma defines the maximal distance, in the new representation
space, between similarly labeled training and testing instances of triplets
allowing us to ensure the respect of the relative constraints on the
testing triplets. By considering that the distance of the projected testing
instances to the projected training instances is lower than a fixed value,
we can define the minimal margin needed between the training triplet elements
such that the testing triplet respects the relative constraint.

Just by bringing those two elements together, we can then show that there
exists a maximum distance in the initial space between similarly labeled
training and testing instances such that the projection of testing triplet
will respect the relative constraints; this maximum distance is related to
the margin of the projection of training triplets.

Let's now consider all our testing instances to be close enough in the
initial space from similarly labeled instances. From previous lemmas, we
know that all training and testing triplets respect the relative
constraints, we can thus define a threshold distinguishing similarly
labeled pair of instances from dissimilarly labeled ones. Thus, the
$\epsilon$-graph obtained based on this threshold now contains only edges
between similarly labeled instances. As our graph is only composed of
connected components of same classes instances, the label propagation will
be optimal.

We also prove that those theoretical element are generalizable for deeper
multi-layered perceptrons.
\section{Experiments}
\label{exp}
After proving theoretically the
interest of our approach for an ideal learned metric, we 
experimentally evaluate the performance of our algorithm.

We evaluate our framework and compare it to different algorithms on several
artificial data sets, composed of 500 instances, chosen for their
increasing degree of complexity. The first artificial data set we will use
is the classical {\em circle} data set. It is obtained by sampling points
on two concentric circles in 2-D, the circle defines the class
membership. The three other artificial data sets are developed variants of
the {\em circle} data set; thus their two first features are generated
through the circle data set generative model. {\em perturbedCircle1} and
{\em perturbedCircle2} data sets both get two additional features,
respectively random and from a 2-D gaussian. {\em perturbedCircle3} is
complemented twice by 2-D gaussian coordinates, depending on the sign of
each of the two first features.  Labels of {\em perturbedCircle2} data sets
are defined depending on the circle on which instances lie and on the
gaussian they are associated to.  For the other artificial data sets, the
label is only dependent on the first two features, as for the initial {\em
circle} data set. We also evaluate our framework on the real data sets {\em
cancer} and {\em vehicule}, which are real data sets of same order of
complexity than the previous data sets; yet {\em vehicule} is composed of 4
classes.

For each data set, we can compute a graph where the 
weights are computed by the the euclidean distance in, respectively, the
initial space and the learned space for LMNN algorithm (\cite{LMNN}), the
ExtraTrees feature selection algorithm (\cite{fsExtra}) and our
approach\footnote{Implemented with torch7 (\cite{Clement2011Torch7})}.
We apply an $\epsilon$-simplification on each graph.
Our network was trained for the euclidean distance
with up to $30 \%$ of the possible triplets from the train set.

\begin{figure}[h!]
\centering
  \begin{minipage}[b][][t]{0.44\linewidth}
  \centering
  \includegraphics[scale=0.21]{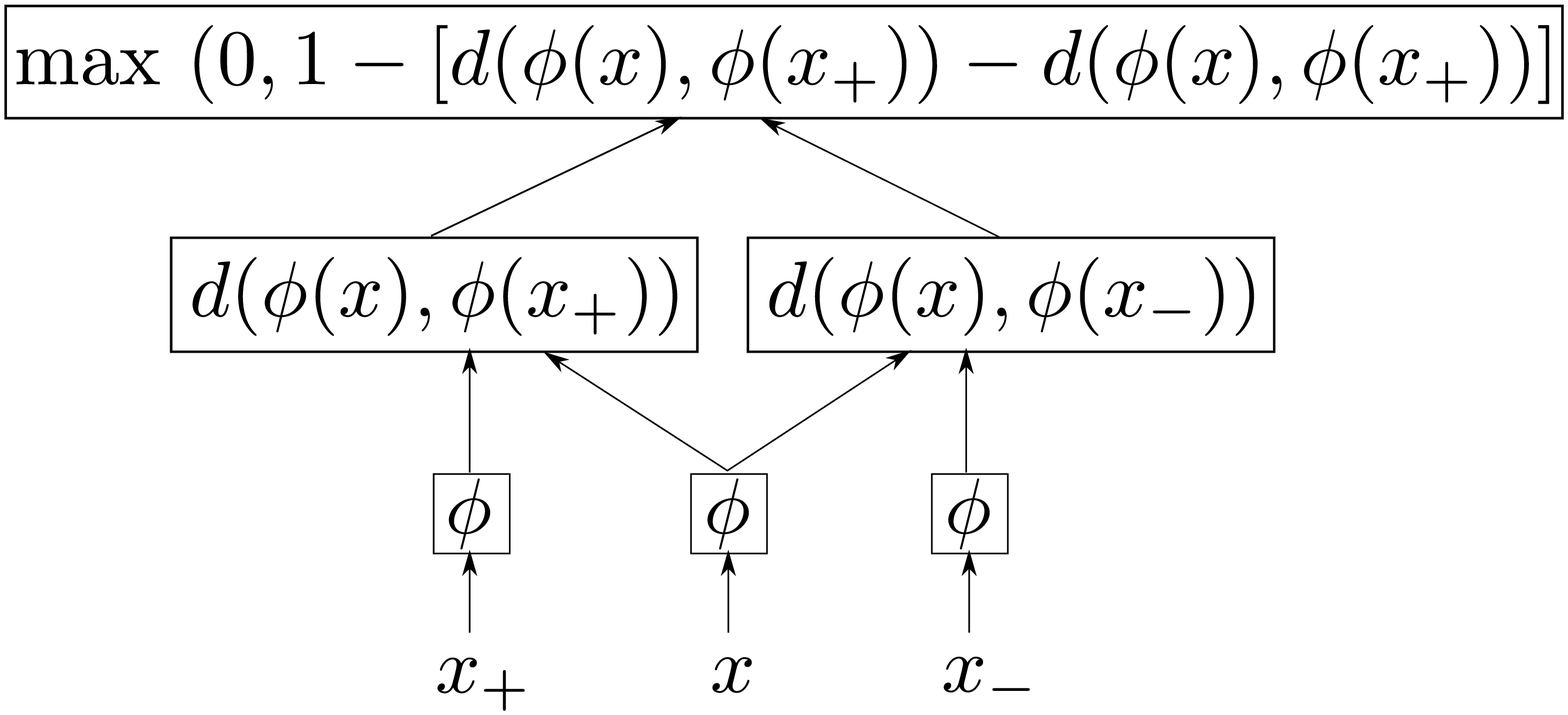}
  \vspace{0.5cm}
  \caption{\label{nnSiamese}Trained siamese network}
  \end{minipage}
  \begin{minipage}[b][][t]{0.55\linewidth}
  \includegraphics[width=0.99\textwidth]{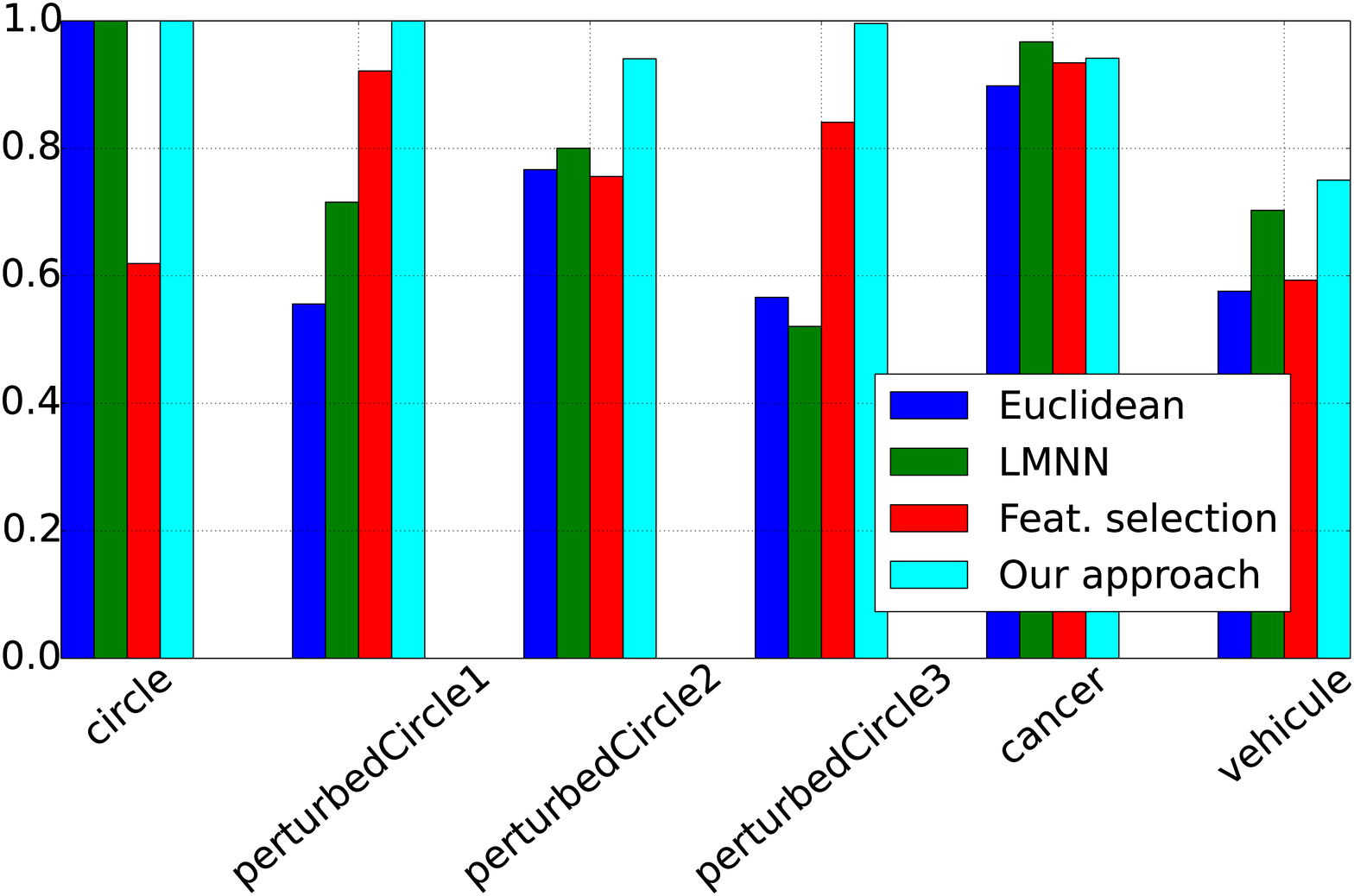}
  \caption{\label{errorDataset}Classification accuracy of label propagation algorithm on $\epsilon$-graph depending on data set }
  \end{minipage}
\end{figure}

In figure \ref{errorDataset}, we compare, for each dataset, the
classification accuracy of the label propagation algorithm performed on the
$\epsilon$-graph computed on the different representation spaces, with 20\% of training instances.  As
expected, the euclidean distance and LMNN perform ideally on the {\em
circle} data set, as the initial space is locally representative of the
labelling similarity; however their performance decrease on artifical data
sets as the initial description space is getting blurred and noisy, like
for {\em perturbedCircle1}, {\em perturbedCircle2} and {\em
perturbedCircle3} data sets.  Feature selection algorithm performs well
enough on {\em perturbedCircle1} and {\em perturbedCircle3} data set, where
the labels are non-perturbed by the added features; however it does not surpass
euclidean distance and LMNN algorithm for other data sets.  On those
artificial data sets, our algorithm performs either ideally or better than
the other algorithms, and was able to learn a more representative projection
space.

Concerning the real dataset {\em cancer}, the different algorithms
perform quit well, broadly similar. Finally, on the {\em vehicule} dataset,
we can see the interest of learning a projection space to obtain a better
label propagation; yet the non-linear projection seems more adapted to the
dataset complexity.

\section{Conclusion} 
\label{conclu}
In this paper, we introduced an algorithm to learn a representation space
for dataset that is adapted to a specific task. We defined and proved a
first theoretical requirement for our algorithm to be optimal; what have
been proved can easily be generalized to multi-layers neural network. 
Experiments on artificial and real data sets confirmed the
relevance of our approach for solving classification task.


%
%
\section{Acknowledgement}
This work was partially supported by a grant from CPER Nord-Pas de
Calais/FEDER DATA Advanced data science and technologies 2015-2020
and ANRT under the grant number CIFRE N\textsuperscript{o} 2013/0961

\bibliography{metricL}
\bibliographystyle{iclr2016_workshop}

\end{document}